\documentclass{article} 
\pdfoutput=1

\usepackage{booktabs}                           
\usepackage{amsmath}                            
\usepackage{amssymb}
\usepackage{amsfonts}                           
\usepackage{nicefrac}                           
\usepackage{microtype}                          
\usepackage{graphicx}                           
\usepackage[font=small,labelfont=bf]{caption}   
\usepackage{bbm}                                
\usepackage{wrapfig}                            
\usepackage{color}                              
\usepackage{iclr2016_conference,times}
\usepackage{hyperref}
\usepackage{tabularx}
\usepackage{url}

\title{
Normalizing the Normalizers:
Comparing and Extending Network Normalization Schemes}

\author{
Mengye Ren\thanks{indicates equal contribution}$^{*\dagger}$, 
Renjie Liao$^{*\dagger}$, Raquel Urtasun$^\dagger$, 
Fabian H. Sinz$^\ddagger$, Richard S. Zemel$^{\dagger\divideontimes}$\\
$^\dagger$University of Toronto, Toronto ON, CANADA\\
$^\ddagger$Baylor College of Medicine, Houston TX, USA\\
$^\divideontimes$Canadian Institute for Advanced Research (CIFAR)\\
\texttt{\{mren, rjliao, urtasun\}@cs.toronto.edu}\\
\texttt{fabian.sinz@epagoge.de, zemel@cs.toronto.edu}
}

\newcommand{\ie}{\textit{i}.\textit{e}.}
\newcommand{\eg}{\textit{e}.\textit{g}.}

\iclrfinalcopy 
\begin{document}

\maketitle

\begin{abstract}
Normalization techniques have only recently begun to be exploited in supervised learning tasks. Batch normalization exploits mini-batch statistics to normalize the activations. This was shown to speed up training and result in better models. However its success has been very limited when dealing with recurrent neural networks. On the other hand, layer normalization normalizes the activations across all activities within a layer. This was shown to work  well in the recurrent setting. In this paper we propose a unified view of  normalization techniques, as forms of divisive normalization, which includes layer and batch normalization as special cases. Our second contribution is the finding that a small modification to these normalization schemes, in conjunction with a sparse regularizer on the activations, leads to significant benefits over standard normalization techniques. We demonstrate the effectiveness of our unified divisive normalization framework  in the context of convolutional neural nets and recurrent neural networks, showing  improvements over baselines in image classification, language modeling as well as super-resolution.  

\end{abstract}

\section{Introduction}

Standard deep neural networks are difficult to train. Even with non-saturating
activation functions such as ReLUs \citep{Krizhevsky2012}, gradient vanishing
or explosion can still occur, since the Jacobian gets multiplied by the input
activation of every layer.  In AlexNet \citep{Krizhevsky2012}, for instance,
the intermediate activations can differ by several orders of magnitude. Tuning
hyperparameters governing weight initialization, learning rates, and various
forms of regularization thus become crucial in optimizing performance.

In current neural networks, normalization abounds.  One technique that has
rapidly become a standard is batch normalization (BN) in which the activations
are normalized by the mean and standard deviation of the training mini-batch
\citep{ioffe2015batch}. At inference time, the activations are normalized by
the mean and standard deviation of the full dataset.  A more recent variant,
layer normalization (LN), utilizes the combined activities of all units within
a layer as the normalizer \citep{ba2016layer}.  Both of these methods have
been shown to ameliorate training difficulties caused by poor initialization,
and help gradient flow in deeper models.

A less-explored form of normalization is divisive normalization (DN)
\citep{Heeger1992a}, in which a neuron's activity is normalized by its
neighbors within a layer. This type of normalization is a well established
canonical computation of the brain \citep{Carandini2011} and has been
extensively studied in computational neuroscience and natural image modelling
(see Section \ref{sec:related}). However, with few exceptions
\citep{Jarrett2009, Krizhevsky2012} it has received little attention in
conventional supervised deep learning.

Here, we provide a unifying view of the different normalization approaches by
characterizing them as the same transformation but along different dimensions
of a tensor, including normalization across examples, layers in the network,
filters in a layer, or instances of a filter response.  We explore the effect
of these varieties of normalizations in conjunction with  regularization, on
the prediction performance compared to baseline models.  The paper thus
provides the first study of divisive normalization in a range of neural
network architectures, including convolutional neural networks (CNNs) and recurrent
neural networks (RNNs), and tasks such as image classification, language
modeling and image super-resolution. We find that DN can achieve results on
par with BN in CNN networks and out-performs it in RNNs and super-resolution,
without having to store batch statistics. We show that casting LN as a form of
DN by incorporating a smoothing  parameter leads to significant gains, in both
CNNs and RNNs. We also find advantages in performance and stability by being
able to  drive learning with higher learning rate in RNNs using DN.  Finally,
we demonstrate that adding an L1 regularizer on the activations before
normalization is beneficial for all forms of normalization.

\section{Related Work}
\label{sec:related}
In this section we first review related work on normalization,
followed by a brief description of regularization in neural networks. 

\subsection{Normalization} Normalization of data prior to training has a long
history in machine  learning. For instance, local contrast normalization used
to be a standard effective tool in vision problems \citep{Pinto2008,
Jarrett2009, Sermanet2012, Le2013}. However, until recently, normalization was
usually not part of the machine learning algorithm itself. Two notable
exceptions are the original AlexNet by \cite{Krizhevsky2012} which includes a
divisive normalization step over a subset of features after ReLU at each pixel
location, and the work by \cite{Jarrett2009} who demonstrated that a
combination of nonlinearities, normalization and pooling improves object
recognition in two-stage networks.

Recently \cite{ioffe2015batch} demonstrated that standardizing the activations
of the summed inputs of neurons over training batches can substantially 
decrease training time in deep neural networks. To avoid covariate shift, 
where the weight gradients in one layer are highly dependent on previous layer 
outputs, Batch Normalization (BN) rescales the summed inputs according to 
their variances under the distribution of the mini-batch data. Specifically, 
if $z_{j,n}$ denotes the activation of a neuron $j$  on example $n$, and $B(n)$
denotes the mini-batch of examples that contains $n$, then BN computes an 
affine function of the activations standardized over each mini-batch:
\[
  \tilde{z}_{n,j} = \gamma
  \frac{z_{n,j} - \mathbb{E}[z_j]}
  {\sqrt{\frac{1}{|B(n)|} (z_{n,j} - \mathbb{E}[z_j])^2}} + \beta
~ ~ ~ \ \ \
\mathbb{E}[z_j] = \frac{1}{|B(n)|} \sum_{m \in B(n)} z_{m,j}
\]
However, training performance in Batch Normalization strongly depends on the 
quality of the aquired statistics and, therefore, the size of the mini-batch. 
Hence, Batch Normalization is harder to apply in cases for which the batch 
sizes are small, such as online learning or data parallelism. While 
classification networks can usually employ relatively larger mini-batches, 
other applications such as image segmentation with convolutional nets use 
smaller batches and suffer from degraded performance. Moreover, application to 
recurrent neural networks (RNNs) is not straightforward and leads to poor 
performance \citep{laurent2015batch}.

Several approaches have been proposed to make Batch Normalization applicable 
to RNNs. \cite{cooijmans2016recurrent} and \cite{Liao2016a} collect separate 
batch statistics for each time step. However, neither of this techniques 
address the problem of small batch sizes and it is unclear how to generalize 
them to unseen time steps.

More recently, \cite{ba2016layer} proposed Layer Normalization (LN), where the 
activations are normalized across all summed inputs within a layer instead of 
within a batch:
\[
  \tilde{z}_{n,j} = \gamma
  \frac{z_{n,j} - \mathbb{E}[z_n]}
  {\sqrt{\frac{1}{|L(j)|} (z_{n,j} - \mathbb{E}[z_n])^2}} + \beta
~ ~ ~ \ \ \
\mathbb{E}[z_n] = \frac{1}{|L(j)|} \sum_{k \in L(j)} z_{n,k}
\]

where $L(j)$ contains all of the units in the same layer as $j$. While
promising results have been shown on RNN benchmarks, direct application of
layer normalization to convolutional layers often leads to a degradation of
performance. The authors hypothesize that since the statistics in
convolutional layers can vary quite a bit spatially, normalization with
statistics from an entire layer might be suboptimal.

\cite{ulyanov2016instancenorm} proposed to normalize each example on spatial
dimensions but not on channel dimension, and was shown to be effective on image style transfer applications \citep{gatys2016styletransfer}.

\cite{Liao2016} proposed to accumulate the normalization statistics over the
entire training phase, and showed that this can speed up training in recurrent
and online learning without a deteriorating effect on the performance. Since
gradients cannot be backpropagated through this normalization operation, the
authors use running statistics of the gradients instead.

Exploring the normalization of weights instead of activations,
\cite{Salimans2016} proposed a reparametrization of the weights into a scale
independent representation and demonstrated that this can speed up training
time.

Divisive Normalization (DN) on the other hand modulates the neural activity by
the activity of a pool of neighboring neurons \citep{Heeger1992a, Bonds1989}.
DN is one of the most well studied and widely found transformations in real
neural systems, and thus has been called a canonical computation of the brain
\citep{Carandini2011}. While the exact form of the transformation can differ,
all formulations model the response of a neuron $\tilde{z}_j$ as a ratio
between the acitivity in a summation field $\mathcal A_j$, and a norm-like
function of the suppression field $\mathcal B_j$
\begin{align}\label{divnorm}
  \tilde{z}_j &= \gamma \frac{\sum_{z_i \in \mathcal{A}_j} u_i z_i} 
                { \left( 
                \sigma^2 + \sum_{z_k \in \mathcal{B}_j} w_k z_k^p 
                \right)^{\frac{1}{p}} },
\end{align}
where $\{u_i\}$ are the summation weights and $\{w_k\}$ the suppression 
weights. 

Previous theoretical studies have outlined several potential computational
roles for divisive normalization such as sensitivity maximization
\citep{Carandini2011}, invariant coding \citep{Olsen2010}, density modelling
\citep{balle2015density}, image compression \citep{Malo2006}, distributed
neural representations \citep{Simoncelli1998}, stimulus decoding
\citep{Ringach2009,Froudarakis2014}, winner-take-all mechanisms
\citep{Busse2009}, attention \citep{Reynolds2009}, redundancy reduction
\citep{Schwartz2001, Sinz2009, Lyu2008a, Sinz2013}, marginalization in neural
probabilistic population codes \citep{Beck2011}, and contextual modulations in
neural populations and perception \citep{Coen-Cagli2015, Schwartz2009}.

\subsection{Regularization}
Various regularization techniques have been applied to neural networks for the purpose of improving generalization and reduce overfitting. They can be roughly divided into two categories, depending on whether they regularize the weights or the activations.

\paragraph{Regularization on Weights:} The most common regularizer on weights
is weight decay which just amounts to using the L2 norm squared of the weight
vector. An L1 regularizer \citep{Goodfellow2016Book} on the weights can also be
adopted to push the learned weights to become sparse.  \cite{Scardapane16}
investigated mixed norms in order to promote group sparsity.

\paragraph{Regularization on Activations:} Sparsity or group sparsity
regularizers on the activations have shown to be effective in the past
\citep{Rozell2008a, Kavukcuoglu2009} and  several regularizers have been
proposed that act directly on the neural activations.  \cite{glorot2011deep}
add a sparse regularizer on the activations after ReLU to encourage sparse
representations.  Dropout developed by \cite{srivastava2014dropout} applies
random masks to the activations in order to discourage them to co-adapt. DeCov
proposed by  \cite{cogswell2015reducing}  tries to minimize the off-diagonal
terms of the sample covariance matrix of activations, thus   encouraging the
activations to be as decorrelated as  possible. \cite{liao2016parsimonious}
utilize a clustering-based regularizer  to encourage the representations to be
compact.

\section{A Unified Framework for Normalizing Neural Nets}
We first compare the three existing forms of normalization, and show that we
can modify batch normalization (BN) and layer normalization (LN) in small ways
to make them have a form that matches divisive normalization (DN).  We present
a general formulation of normalization, where existing normalizations
involve alternative schemes of accumulating information. Finally, we propose a
regularization term that can be optimized jointly with these normalization
schemes to encourage decorrelation and/or improve generalization performance.

\subsection{General Form of Normalization}
Without loss of generality, we denote the hidden input activation of one
arbitrary layer in a deep neural network as  $\mathbf{z} \in \mathbb{R}^{N
\times L}$. Here $N$ is the mini-batch size. In the case of a CNN, $L=H \times
W \times C$, where  $H, W$ are the height and width of the convolutional
feature map and $C$ is the number of filters. For an RNN or fully-connected
layers of a neural net, $L$ is the number of hidden units.

\begin{figure}
\centering 
\includegraphics[width=0.7\textwidth]{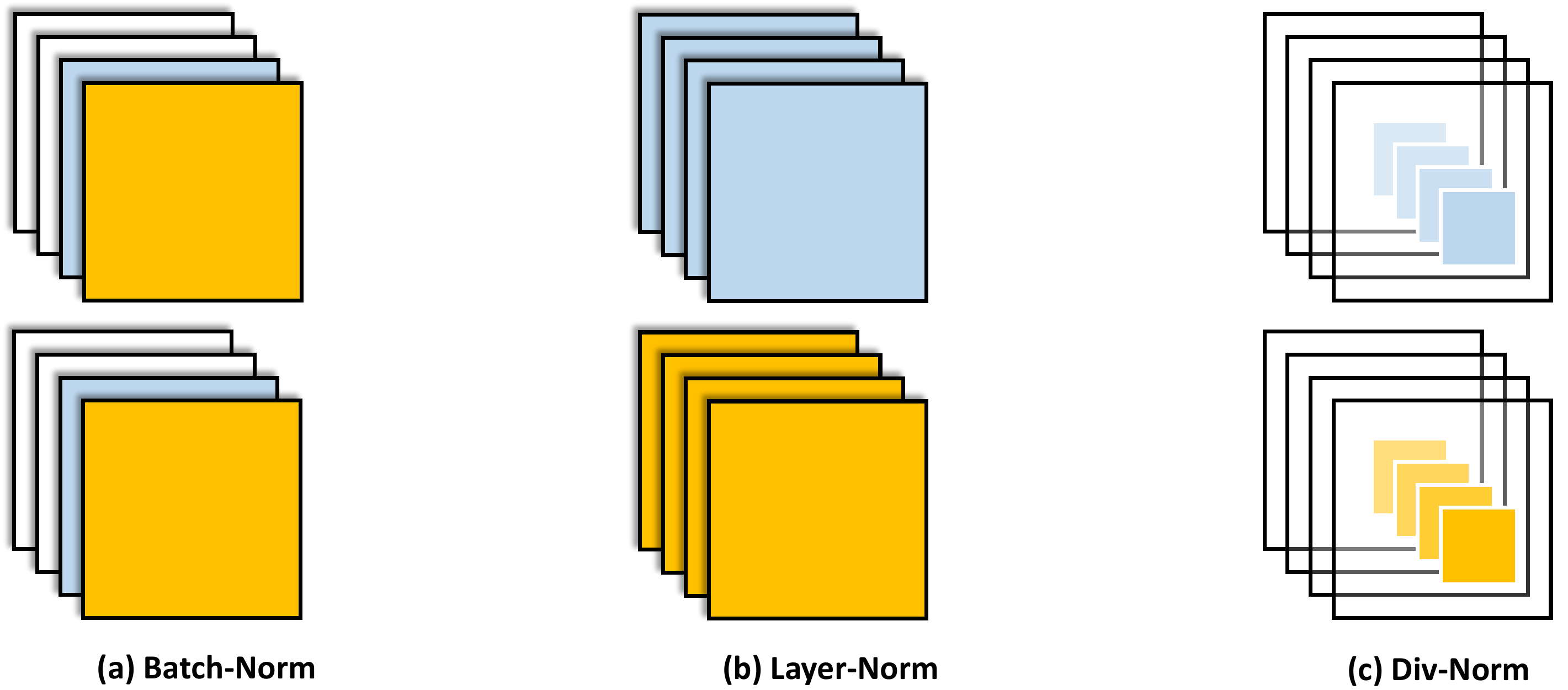}
\caption{Illustration of different normalization schemes, in a CNN. Each $H 
\times W$-sized feature map is depicted as a rectangle; overlays depict 
instances in the set of $C$ filters; and two examples from a mini-batch of 
size $N$ are shown, one above the other. The colors show the 
summation/suppression fields of each scheme.}
\label{fig_norm}
\end{figure}

Different normalization methods gather statistics from different ranges
of the tensor and then perform normalization. Consider the following general 
form:
\begin{align}
\label{general_form}
z_{n, j}  &= \sum_i w_{i, j} x_{n, i} + b_j \\
v_{n, j}  &= z_{n, j} - \mathbb{E}_{\mathcal{A}_{n, j}}[z]\\
\tilde{z}_{n, j} &= \frac{ v_{n, j}}  {\sqrt{ \sigma^2 + \mathbb{E}_{\mathcal{B}_{n, j}}[v^2]}}
\end{align}
where $\mathcal{A}_j$ and $\mathcal{B}_j$ are subsets of $z$ and $v$
respectively. $\mathcal{A}$ and $\mathcal{B}$ in standard divisive
normalization are referred to as summation and suppression fields
\citep{Carandini2011}. One can cast each normalization scheme into this
general formulation, where the schemes vary based on how they define these two
fields. These definitions are specified in Table \ref{tbl_norm_range}. 
Optional parameters $\gamma$ and $\beta$ can be added in the 
form of $\gamma_j \tilde{z}_{n, j} + \beta_j$ to increase the degree of 
freedom.

\begin{table}[h]
\begin{center}
\def\arraystretch{1.5}
\begin{tabular}{c | c| c }
Model 		& Range & Normalizer Bias\\ 
\hline 
BN 	&
\begin{tabular}{l} 
$\mathcal{A}_{n,j}=\{z_{m,j}: m \in [1,N],j \in [1,H]\times[1, W]\}$ \\ 
$\mathcal{B}_{n,j}=\{v_{m,j}: m \in [1,N],j \in [1,H]\times[1, W]\}$
\end{tabular} & $\sigma=0$ \\
LN   	&
\begin{tabular}{ll}
$\mathcal{A}_{n,j}=\{z_{n,i}: i \in [1,L]\}$ & 
$\mathcal{B}_{n,j}=\{v_{n,i}: i \in [1,L]\}$
\end{tabular} & $\sigma=0$ \\
DN 		&
\begin{tabular}{ll}
$\mathcal{A}_{n,j}=\{z_{n,i}: d(i,j) \le R_{\mathcal{A}}\}$ & 
$\mathcal{B}_{n,j}=\{v_{n,i}: d(i,j) \le R_{\mathcal{B}}\}$
\end{tabular} & $\sigma\ge 0$ \\
\end{tabular}
\caption{Different choices of the summation and suppression fields $\mathcal{A}$ and 
$\mathcal{B}$, as well as the constant $\sigma$ in the normalizer lead to known normalization schemes in neural networks. $d(i, j)$ denotes an arbitrary 
distance between two hidden units $i$ and $j$, and $R$ denotes the neighbourhood radius.
}
\label{tbl_norm_range}
\end{center}
\end{table}

Fig.~\ref{fig_norm} shows a visualization of the normalization field in a 
4-D ConvNet tensor setting. Divisive normalization happens within a local 
spatial window of neurons across filter channels. Here we set $d(\cdot, \cdot)$
to be the spatial $L_\infty$ distance.

\subsection{New Model Components}
\paragraph{Smoothing the Normalizers:} One obvious way in which the
normalization schemes differ is in terms of the information that they combine
for normalizing the activations.  A second more subtle but important
difference between standard BN and LN as opposed to DN is the smoothing term
$\sigma$, in the denominator of Eq. (\ref{divnorm}). This term allows some
control of the bias of the variance estimation, effectively smoothing the
estimate.  This is beneficial because divisive normalization does not utilize
information from the mini-batch as in BN, and combines information from a
smaller field than LN. A similar but different denominator bias term
$\max(\sigma, c)$ appears in \citep{Jarrett2009}, which is
active when the activation variance is small.
However, the clipping function makes the transformation not invertible, losing
scale information. 

\begin{figure}
\centering 
\includegraphics[width=0.44\textwidth,trim={0cm 0cm 0cm 0.25cm},clip]{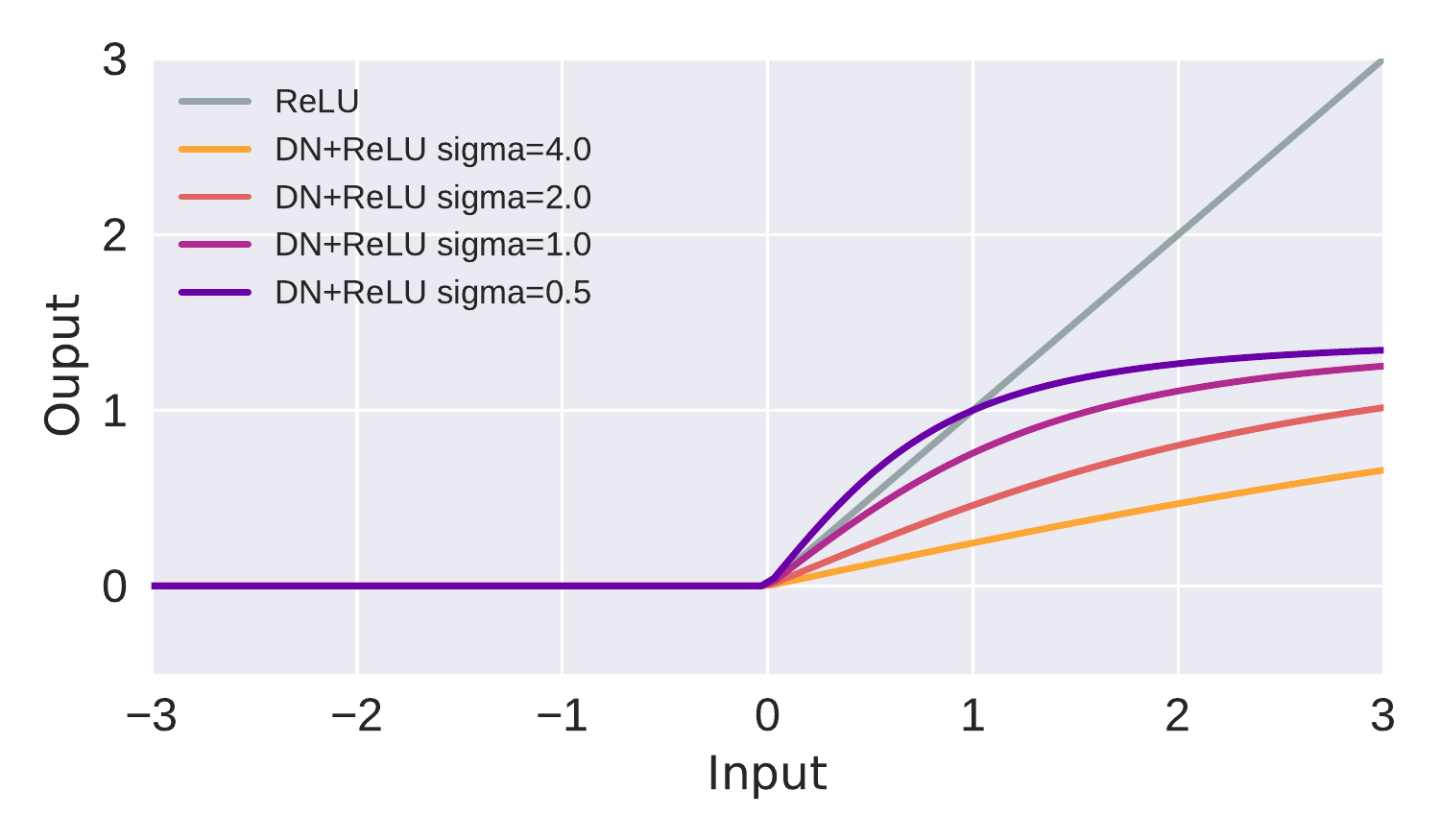}
\includegraphics[width=0.54\textwidth]{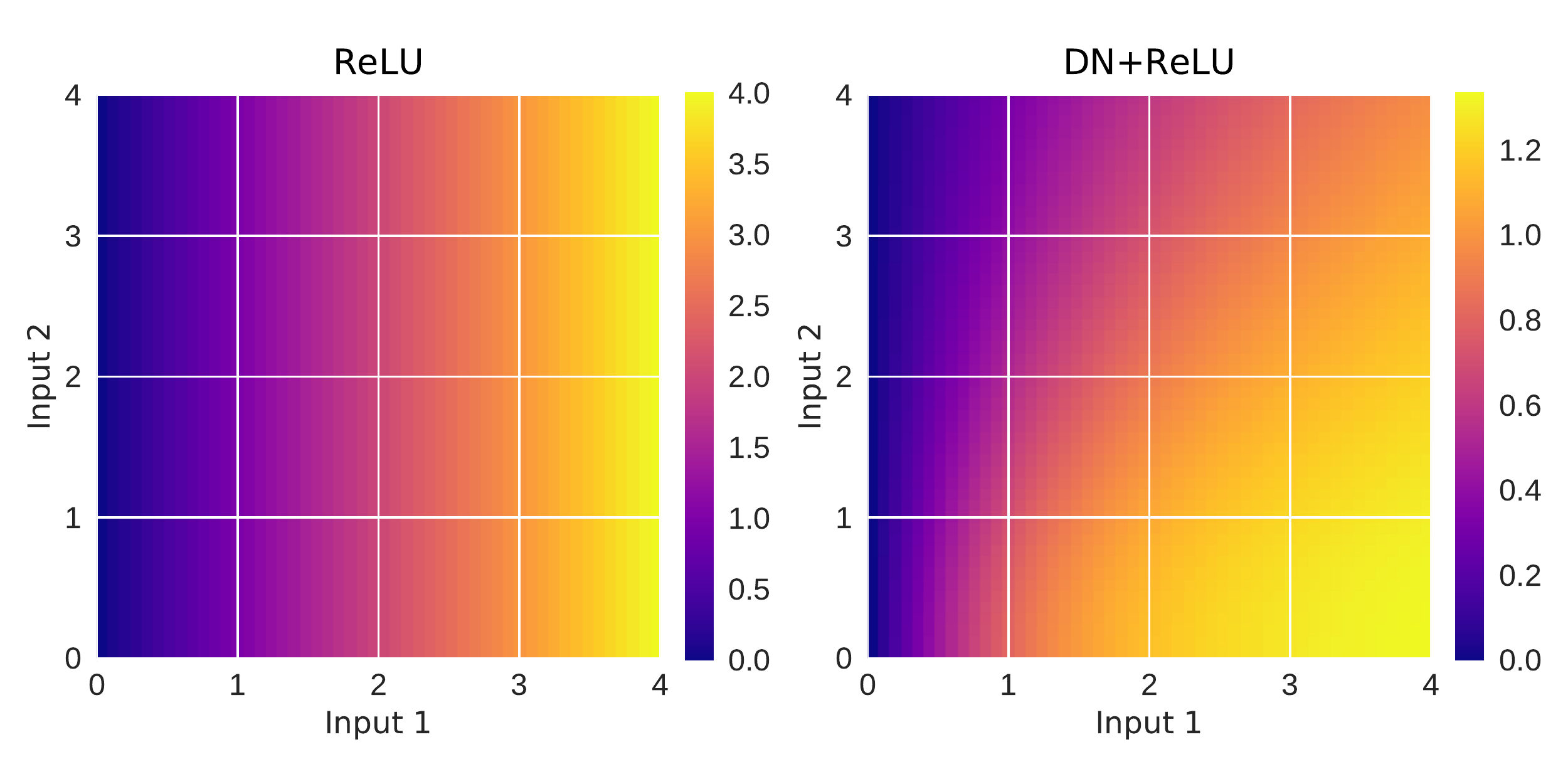}
\caption{
Divisive normalization followed by ReLU can be viewed as a new activation
function. Left: Effect of varying $\sigma$ in this activation function.
Right: Two units affect each other's activation in the DN+ReLU
formulation.}
\label{fig_act_fn}
\end{figure}

Moreover, if we take the nonlinear activation function after normalization
into consideration, we find that $\sigma$ will change the overall properties
of the non-linearity. To illustrate this effect, we use a simple 1-layer
network which consists of: two input units, one divisive normalization
operator, followed by a ReLU activation function.  If we fix one input unit to
be 0.5, varying the other one with different values of $\sigma$ produces
different output curves  (Fig. \ref{fig_act_fn}, left).  These curves exhibit
different non-linear properties compared to the standard ReLU.  Allowing the
other input unit to vary as well results in different activation functions of
the  first unit depending on the activity of the second  (Fig.
\ref{fig_act_fn}, right). This illustrates potential benefits of including
this smoothing term $\sigma$, as it effectively modulates the rectified
response to vary from a linear to a highly saturated response.

In this paper we propose modifications of the standard BN and LN which borrow this additive term $\sigma$ in the denominator from DN.  We study the effect of incorporating this smoother in the respective normalization schemes below.

\paragraph{L1 regularizer:} Filter responses on lower layers in deep neural
networks can be quite correlated which might impair the estimate of the
variance in the normalizer. More independent representations help disentangle
latent factors and boost the networks performance \citep{Higgins2016}.
Empirically, we found that putting a sparse (L1) regularizer  \begin{equation}
\label{eq:l1} \mathcal{L}_{L1} = \alpha \frac{1}{N L} \sum_{n, j} |v_{n, j}|
\end{equation}  on the centered activations $v_{n,j}$ helps decorrelate the
filter responses (Fig.~\ref{fig:joint_histo}). Here, $N$
is the batch size and $L$ is the number of hidden units, and
$\mathcal{L}_{L1}$ is the regularization loss which is added to the training
loss.

A possible explanation for this effect is that the L1 regularizer might have a
similar effect as maximum likelihood estimation of an independent Laplace
distribution. To see that, let $ p_v\left(\mathbf{v}\right)
\propto\exp\left(-\left\Vert \mathbf{v}\right\Vert _{1}\right) $ and
$\mathbf{x}=W^{-1}\mathbf{v}$, with $W$ a full rank invertible matrix. Under this
model  $p_x\left(\mathbf{x}\right)  =p_v\left(W\mathbf x\right)\left|\det
W\right|$. Then, minimization of the L1 norm of the activations under the
volume-conserving constraint $\det A=\text{const.}$ corresponds to maximum
likelihood on that model, which would encourage decorrelated responses. We do
not enforce such a constraint, and the filter matrix might even not be
invertible. However, the supervised loss function of the network benefits from
having diverse non-zero filters. This encourages the network to not collapse
filters along the same direction or put them to zero, and might act as a
relaxation of the volume-conserving constraint.

\subsection{Summary of new models}
{\bf DN and DN*:} We propose DN as a new local normalization scheme in neural
networks. In convolutional layers, it operates on a local spatial window
across filter  channels, and in fully connected layers it operates on a slice
of a hidden  state vector. Additionally, DN* has a L1 regularizer on the
pre-normalization centered activation ($v_{n, j}$).

{\bf BN-s and BN*:} To compare with DN and DN*, we also propose modifications
to original BN: we denote BN-s with $\sigma^2$ in the denominator's square
root, and BN* with the L1 regularizer on top of BN-s.

{\bf LN-s and LN*:} We apply the same changes as from BN to BN-s and BN*. In
order to narrow the differences in the normalization schemes down to a few
parameter choices, we additionally  remove the affine transformation
parameters $\gamma$ and $\beta$ from LN such  that the difference between LN*
and DN* is only the size of the normalization  field.  $\gamma$ and $\beta$
can really be seen as a separate layer and in practice  we find that they do
not improve the performance in the presence of $\sigma^2$.

\section{Experiments}

We evaluate the normalization schemes on three different tasks:
\begin{itemize}
    \item {\bf CNN image classification:} We apply different normalizations on CNNs trained 
    on the CIFAR-10/100 datasets for image recognition, each of
    which contains 50,000 training images and 10,000 test images. Each 
    image is
    of size 32 $\times$ 32 $\times$ 3 and has been labeled an object class out 
    of 10 or 100 total number of classes.
    \item {\bf RNN language modeling:} We  apply different normalizations on RNNs
    trained on the Penn Treebank dataset for language modeling, containing 42,068 
    training sentences, 3,370 validation sentences, and 3,761 test sentences.
    \item {\bf CNN image super-resolution:} We train a CNN on low resolution images and
    learn cascades of non-linear filters to smooth the upsampled images. We 
    report performance of trained CNN on the standard Set 14 and 
    Berkeley 200 dataset.
\end{itemize}

For each model, we perform a grid search of three or four choices of each 
hyperparameter including the smoothing constant $\sigma$, and L1
regularization constant $\alpha$, and learning rate $\epsilon$ on the 
validation set.

\subsection{CIFAR Experiments}
We used the standard CNN model provided in the Caffe library. The 
architecture is summarized in Table~\ref{tab:cifar-cnn-table}. 
We apply normalization before each ReLU function. We implement 
DN as a convolutional operator, fixing the local window size to 
$5 \times 5$, $3 \times 3$, $3 \times 3$
for the three convolutional layers in all the CIFAR experiments.

\begin{table}[t]
\caption{CIFAR CNN specification}
\label{tab:cifar-cnn-table}
\begin{center}
\begin{tabular}{l|l|l|l}
{\bf Type}  & {\bf Size}        &  {\bf Kernel}            & {\bf Stride} \\
\hline\hline                                                                  
input      &$32\times32\times3 $& -                        & -            \\
conv +relu &$32\times32\times32$&$5\times5\times3\times32$ & 1            \\
max pool   &$16\times16\times32$&$3\times3$                & 2            \\
conv +relu &$16\times16\times32$&$5\times5\times32\times32$& 1            \\
avg pool   &$8 \times8 \times32$&$3\times3$                & 2            \\
conv +relu &$8 \times8 \times64$&$5\times5\times32\times64$& 1            \\
avg pool   &$4 \times4 \times64$&$3\times3$                & 2            \\
fully conn. linear & $64$       &  -                       & -            \\
fully conn. linear & $10$ or $100$ &  -                    & -           
\end{tabular}
\end{center}
\end{table}

We set the learning rate to 1e-3 and momentum 0.9 for all experiments. The 
learning rate schedule is set to \{5K, 30K, 50K\} for the baseline model and 
to \{30K, 50K, 80K\} for all other models. At every stage we multiply the 
learning rate by 0.1. Weights are randomly initialized from a zero-mean normal 
distribution with standard deviation \{1e-4, 1e-2, 1e-2\} for the 
convolutional layers, and \{1e-1, 1e-1\} for fully connected layers. Input 
images are centered on the dataset image mean.

Table~\ref{tab:cifar-results} summarizes the test performances of 
BN*, LN* and DN*, compared to the performance of a few baseline models and the 
standard batch and layer normalizations. We also add standard regularizers to 
the baseline model: L2 weight decay (WD) and dropout. Adding the smoothing 
constant and L1 regularization consistently improves the classification 
performance, especially for the original LN. The modification of LN makes it 
now better than the original BN, and only slightly worse than BN*. DN*
achieves comparable performance to BN* on both datasets, but only relying on a 
local neighborhood of hidden units.

\begin{table}
\caption{CIFAR-10/100 experiments}
\label{tab:cifar-results}
\begin{center}
\begin{tabular}{l|ll}
{\bf Model} & {\bf CIFAR-10 Acc.} & {\bf CIFAR-100 Acc.}\\
\hline\hline
Baseline        & 0.7565        & 0.4409          \\
Baseline +WD +Dropout & 0.7795        & 0.4179          \\
BN              & 0.7807        & 0.4814          \\
LN              & 0.7211        & 0.4249          \\
\hline
BN*             & {\bf 0.8179}  & {\bf 0.5156}    \\
LN*             & 0.8091        & 0.4957          \\
DN*             & 0.8122        & 0.5066          \\
\end{tabular}
\end{center}
\end{table}

\begin{wrapfigure}{r}{5cm}
\centering
\includegraphics[height=6cm,trim={0.5cm 0cm 2.8cm 0.0cm},clip]{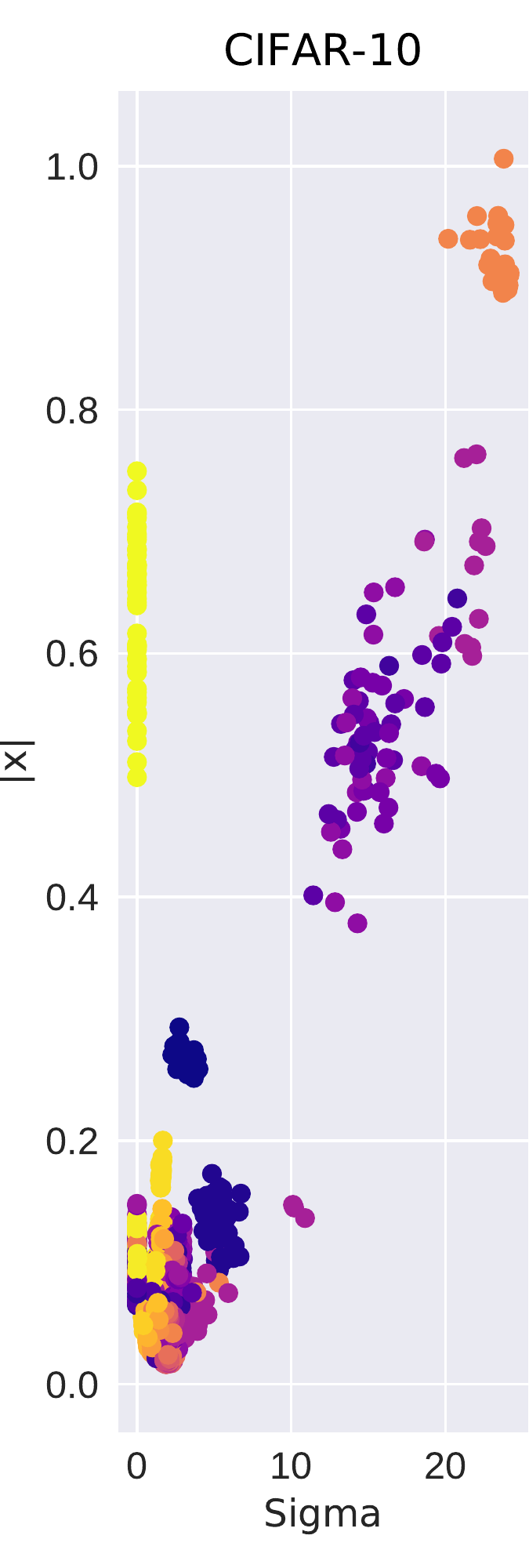}
\includegraphics[height=6cm,trim={1.5cm 0cm 0cm 0.0cm},clip]{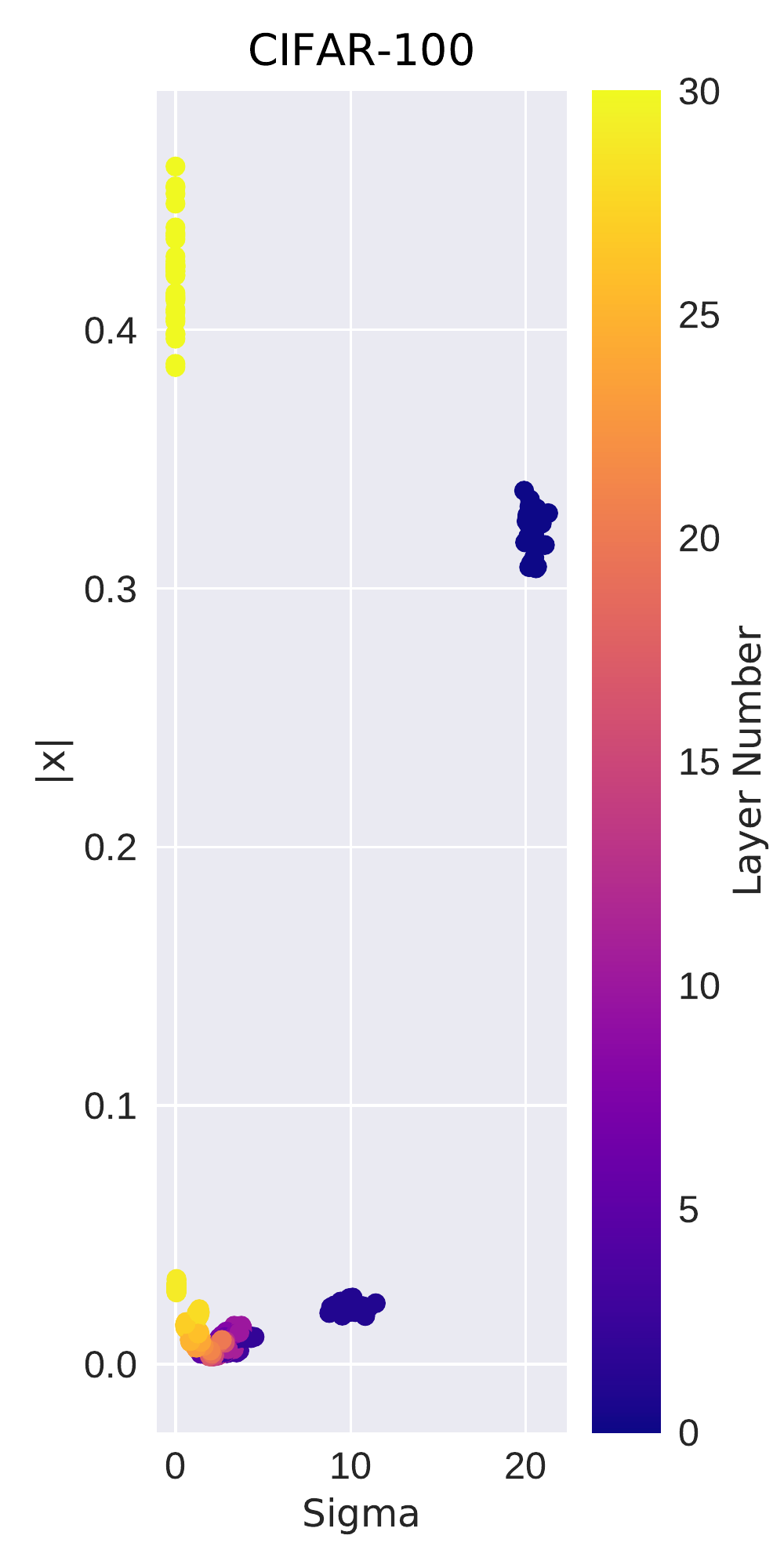}
\caption{Input scale ($|x|$) vs. learned $\sigma$ at each layer, color coded by the layer number in ResNet-32, trained on CIFAR-10 (left), and CIFAR-100 (right).}
\label{fig:sigma_resnet}
\end{wrapfigure}

\paragraph{ResNet Experiments.} Residual networks (ResNet)
\citep{he2016resnet}, a type of CNN with residual connections between
layers, achieve impressive performance on many image classification
benchmarks. The original architecture uses BN by default. If we remove BN,
the architecture is very difficult to train or converges to a poor  solution.
We first reproduced the original BN ResNet-32, obtaining 92.6\% accuracy
on CIFAR-10, and 69.8\% on CIFAR-100. Our best DN model achieves 91.3\% and 66.6\%,
respectively. While this performance is lower than the original BN-ResNet,
there is certainly room to improve as we have not performed any hyperparameter
optimization. Importantly, the beneficial effects of sigma (2.5\% gain on CIFAR-100) and
the L1 regularizer (0.5\%) are still found, even in the presence of other
regularization techniques such as data augmentation and weight decay
in the training.

Since the number of sigma hyperparameters scales with the number of layers, we
found that setting sigma as a learnable parameter for each layer helps the
performance (1.3\% gain on CIFAR-100). Note that training this parameter is
not possible in the formulation by \cite{Jarrett2009}. The learned sigma shows
a clear trend: it tends to decrease with depth, and in the last convolution
layer it approaches 0 (see Fig.~\ref{fig:sigma_resnet}).

\subsection{RNN experiments}

To apply divisive normalization in fully connected layers of RNNs, we consider 
a local neighborhood in the 
hidden state vector $\mathbf{h}_{j-R: j+R}$, where $R$ is the radius of the 
neighborhood. Although the hidden states are randomly initialized, this structure
will impose local competition among the neighbors.
\begin{align}
v_j  &= z_j - \frac{1}{2R + 1} \sum_{r=-R}^R z_{j + r} \\
\tilde{z}_j &= \frac{ v_j  } { \sqrt{ \sigma^2 +  \frac{1}{2R + 1}
\sum_{r=-R}^R v_{j+r}^2  } }
\end{align}
We follow \cite{cooijmans2016recurrent}'s batch 
normalization implementation for RNNs: 
normalizers are separate for input transformation and hidden transformation.
Let $BN(\cdot)$, $LN(\cdot)$, $DN(\cdot)$ be BatchNorm, LayerNorm and DivNorm, 
and $g$ be either tanh or ReLU.

\begin{align}
\mathbf{h}_{t+1}       &=g(W_x\mathbf{x}_t+W_h\mathbf{h}_{t-1} + b)\\
\mathbf{h}_{t+1}^{(BN)}&=g(BN(W_x\mathbf{x}_t+b_x)+BN(W_h\mathbf{h}_{t-1}^{(BN)}+b_h))\\
\mathbf{h}_{t+1}^{(LN)}&=g(LN(W_x\mathbf{x}_t+W_h\mathbf{h}_{t-1}^{(LN)}+b))\\
\mathbf{h}_{t+1}^{(DN)}&=g(DN(W_x\mathbf{x}_t+W_h\mathbf{h}_{t-1}^{(DN)}+b))
\end{align}

Note that in recurrent BN, the additional parameters $\gamma$ and $\beta$ are
shared across timesteps whereas the moving averages of batch  statistics are
not shared. For the LSTM version, we followed the released  implementation
from the authors of layer normalization
\footnote{\url{https://github.com/ryankiros/layer-norm}},  and apply LN at the
same places as BN and BN*, which is after the  linear transformation of $W_x
\mathbf{x}$ and $W_h \mathbf{h}$ individually.  For LN* and DN, we modified
the places of normalization to be at each  non-linearity, instead of jointly
with a concatenated vector for different  non-linearity. We found that  this
modification improves the performance and makes the formulation clearer since
normalization is always a combined operation with the activation function. We
include details of the LSTM implementation in the Appendix.

The RNN model is provided by the Tensorflow library \citep{tensorflow} and
the LSTM version was originally proposed in  \cite{zeremba2014rnnreg}.  We
used a two-layer stack-RNN of size 400 (vanilla RNN) or 200 (LSTM). $R$ is set
to 60 (vanilla RNN) and 30 (LSTM). We tried both tanh and ReLU as the
activation function for the vanilla RNN. For unnormalized baselines and
BN+ReLU, the initial  learning rate is set to 0.1 and decays by half every
epoch, starting at the 5th  epoch for a maximum of 13 epochs. For the other
normalized models, the initial  learning rate is set to 1.0 while the schedule
is kept the same. Standard stochastic gradient descent is used in all RNN
experiments, with gradient clipping at 5.0.

Table~\ref{tab:ptb_results} shows the test set perplexity for LSTM models and
vanilla models.  Perplexity is defined as $\text{ppl} = \exp(-\sum_x \log
p(x))$.  We find that BN and LN alone do not improve the final performance
relative to the  baseline, but similar to what we see in the CNN experiments,
our modified versions BN* and LN* show significant improvements.  BN* on RNN
is outperformed by both LN* and DN.  By applying our normalization, we can
improve the vanilla RNN perplexity by 20\%, comparable to an LSTM baseline
with the same hidden dimension.

\begin{table}
\caption{PTB Word-level language modeling experiments}
\label{tab:ptb_results}
\begin{center}
\begin{tabular}{l|lll}
{\bf Model}   & {\bf LSTM}    & {\bf TanH RNN}& {\bf ReLU RNN}     \\
\hline \hline                                                                         
Baseline      & 115.720       & 149.357       & 147.630            \\
BN            & 123.245       & 148.052       & 164.977            \\
LN            & 119.247       & 154.324       & 149.128            \\
\hline                                                               
BN*           & 116.920       & 129.155       & 138.947            \\
LN*           & {\bf 101.725} & 129.823       & {\bf 116.609}      \\
DN*           & 102.238       & {\bf 123.652} & 117.868            \\
\end{tabular}
\end{center}
\end{table}

\subsection{Super Resolution Experiments}
We also evaluate DN on the low-level computer vision problem of single image
super-resolution. We adopt the SRCNN model of \cite{dong2016image} as the
baseline which consists of 3 convolutional layers and 2 ReLUs. From bottom to
top layers, the sizes of the filters are 9, 5, and 5 \footnote{We use the
setting of the best model out of all three SRCNN candidates.}. The number of
filters are 64, 32, and 1, respectively. All the filters are initialized with
zero-mean Gaussian and standard deviation 1e-3. Then we respectively apply
batch normalization (BN) and our divisive normalization with L1 regularization
(DN*) to the convolutional feature maps before ReLUs. We construct the
training set in a similar manner as \cite{dong2016image} by randomly cropping
5 million patches (size $33 \times 33$) from a subset of the ImageNet dataset
of \cite{deng2009imagenet}. We only train our model for 4 million iterations
which is less than the one adopted by SRCNN, \ie, 15 million, as the gain of
PSNR and SSIM by spending that long time is marginal.

\begin{figure}[t]
\begin{minipage}[t]{\linewidth}
\begin{minipage}[t]{0.24\linewidth}
\centering
\includegraphics[width=1.0\linewidth]{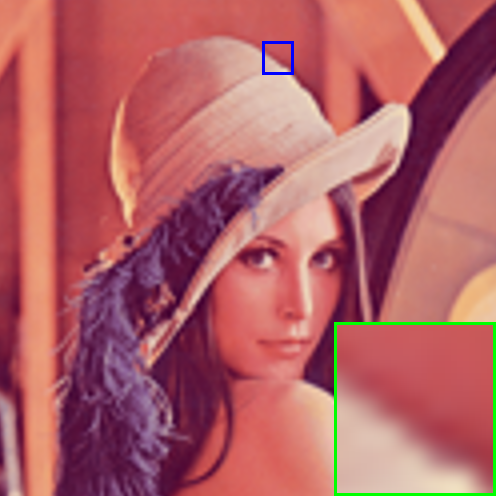}
{PSNR 29.84dB}
\end{minipage}
\begin{minipage}[t]{0.24\linewidth}
\centering
\includegraphics[width=1.0\linewidth]{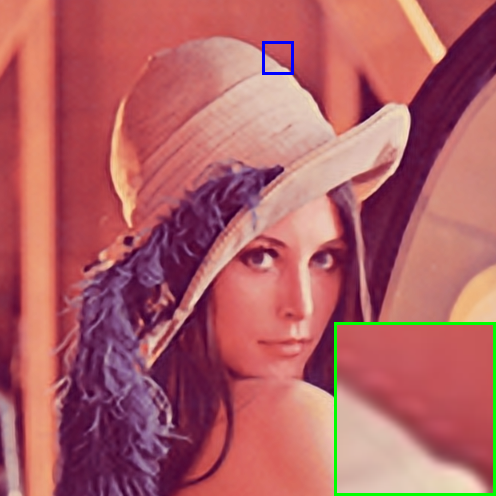}
{PSNR 31.33dB}
\end{minipage}
\begin{minipage}[t]{0.24\linewidth}
\centering
\includegraphics[width=1.0\linewidth]{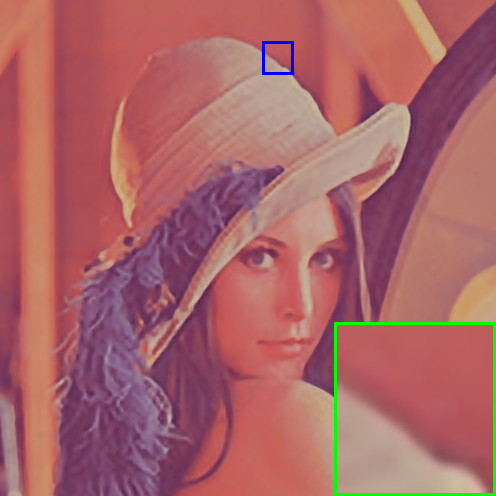}
{PSNR 23.94dB}
\end{minipage}
\begin{minipage}[t]{0.24\linewidth}
\centering
\includegraphics[width=1.0\linewidth]{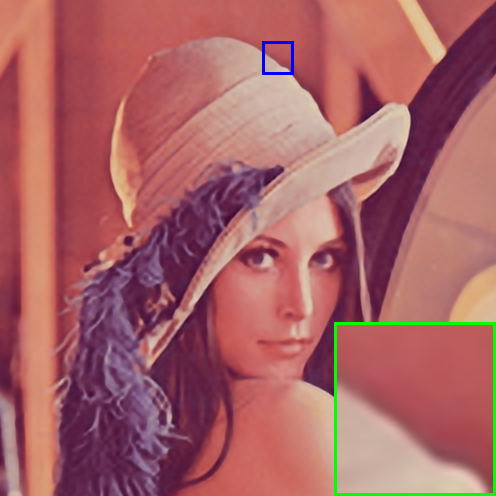}
{PSNR 31.46dB}
\end{minipage}
\end{minipage}
\begin{minipage}[t]{\linewidth}
\begin{minipage}[t]{0.24\linewidth}
\centering
\includegraphics[width=1.0\linewidth]{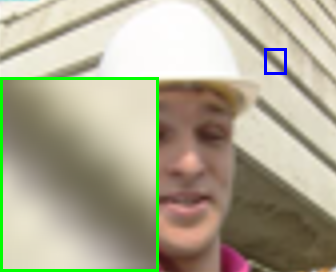}
{PSNR 29.41dB}
\end{minipage}
\begin{minipage}[t]{0.24\linewidth}
\centering
\includegraphics[width=1.0\linewidth]{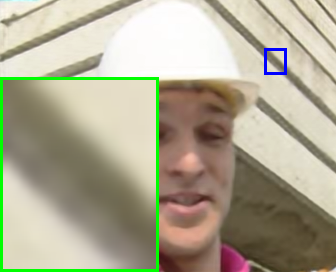}
{PSNR 33.14dB}
\end{minipage}
\begin{minipage}[t]{0.24\linewidth}
\centering
\includegraphics[width=1.0\linewidth]{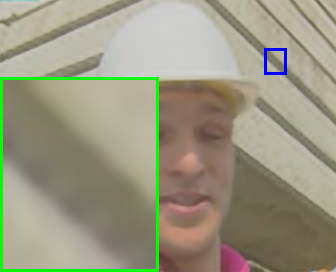}
{PSNR 21.88dB}
\end{minipage}
\begin{minipage}[t]{0.24\linewidth}
\centering
\includegraphics[width=1.0\linewidth]{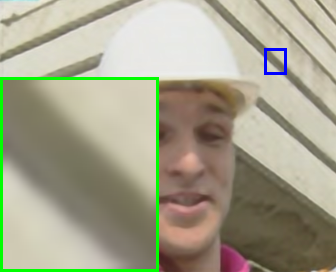}
{PSNR 33.43dB}
\end{minipage}
\end{minipage}
\begin{minipage}[t]{\linewidth}
\begin{minipage}[t]{0.24\linewidth}
\centering
\includegraphics[width=1.0\linewidth]{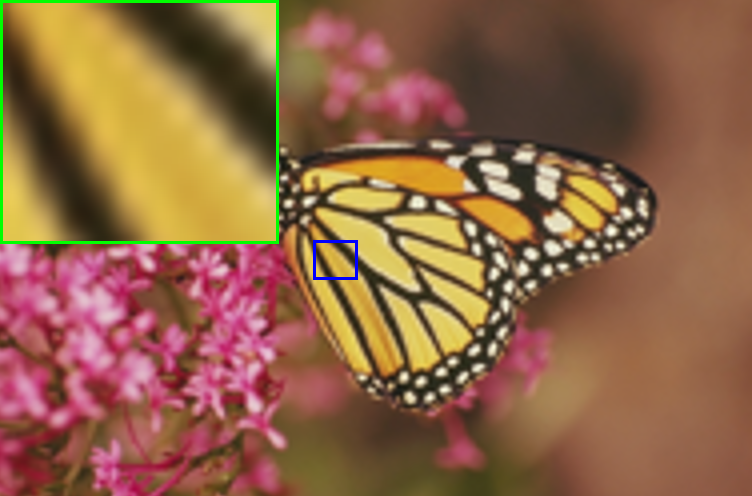}
{PSNR 27.46dB \\ (a) Bicubic}
\end{minipage}
\begin{minipage}[t]{0.24\linewidth}
\centering
\includegraphics[width=1.0\linewidth]{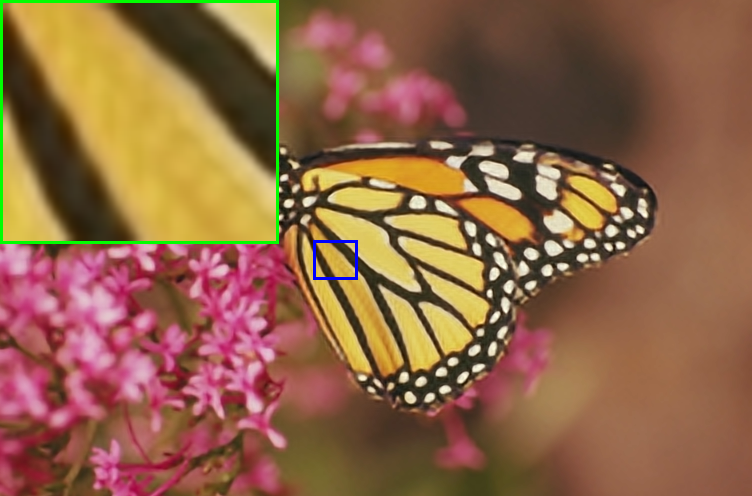}
{PSNR 30.12dB \\ (b) SRCNN}
\end{minipage}
\begin{minipage}[t]{0.24\linewidth}
\centering
\includegraphics[width=1.0\linewidth]{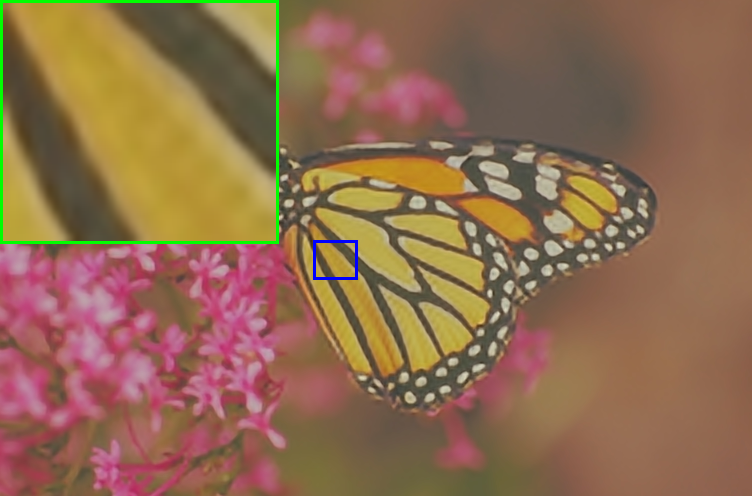}
{PSNR 23.91dB \\ (c) BN}
\end{minipage}
\begin{minipage}[t]{0.24\linewidth}
\centering
\includegraphics[width=1.0\linewidth]{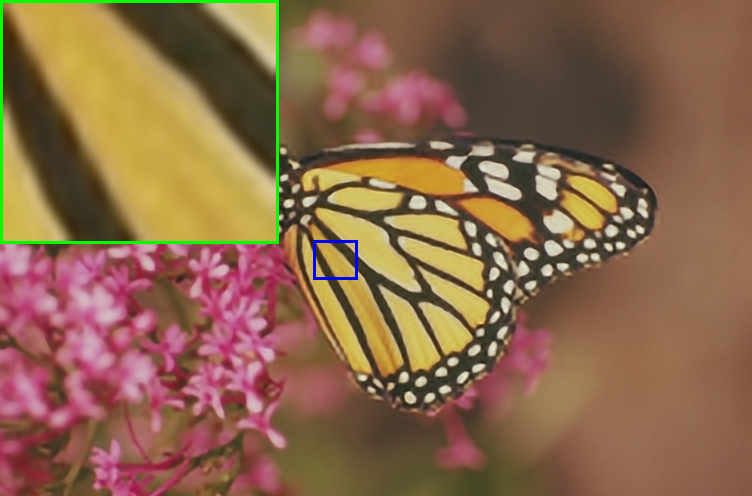}
{PSNR 30.19dB \\ (d) DN*}
\end{minipage}
\end{minipage}
\caption{Comparisons at a magnification factor of 4.}
\label{fig_single_SR_1}
\end{figure}

We report the average test results, utilizing the standard metrics PSNR and
SSIM \citep{wang2004image}, on two standard test datasets Set14
\citep{zeyde2010single} and BSD200 \citep{martin2001database}. We compare with
two state-of-the-art single image super-resolution methods, A+
\citep{timofte2013anchored} and SRCNN \citep{dong2016image}. All measures are
computed on the Y channel of YCbCr color space. We also provide a visual
comparison in Fig. \ref{fig_single_SR_1}.

\begin{table}[t]
\begin{center}
\caption{Average test results of PSNR and SSIM on Set14 Dataset.}
\begin{tabular}{l|llll}
\textbf{Model} & \textbf{PSNR} (x3) & \textbf{SSIM} (x3) & \textbf{PSNR} (x4) & \textbf{SSIM} (x4) \\
\hline \hline
Bicubic	& 27.54 & 0.7733 & 26.01 & 0.7018 \\
A+			& 29.13 & 0.8188 & 27.32 & 0.7491 \\
SRCNN   & 29.35 & 0.8212 & 27.53 & 0.7512 \\ 
BN   		& 22.31 & 0.7530 & 21.40 & 0.6851 \\
DN* 		& \textbf{29.38} & \textbf{0.8229} & \textbf{27.64} & \textbf{0.7562} \\
\end{tabular}
\label{tbl_sr_1}
\end{center}
\end{table}

\begin{table}[t]
\begin{center}
\caption{Average test results of PSNR and SSIM on BSD200 Dataset.}
\begin{tabular}{l|llll}
\textbf{Model} & \textbf{PSNR} (x3) & \textbf{SSIM} (x3) & \textbf{PSNR} (x4) & \textbf{SSIM} (x4) \\
\hline \hline
Bicubic & 27.19 & 0.7636 & 25.92 & 0.6952 \\
A+			& 27.05 & 0.7945 & 25.51 & 0.7171 \\
SRCNN   & 28.42 & 0.8100 & 26.87 & 0.7378 \\
BN   		& 21.89 & 0.7553 & 21.53 & 0.6741 \\
DN* & \textbf{28.44} & \textbf{0.8110} & \textbf{26.96} & \textbf{0.7428} \\
\end{tabular}
\label{tbl_sr_2}
\end{center}
\end{table}

As show in Tables \ref{tbl_sr_1} and \ref{tbl_sr_2} DN* outperforms the strong
competitor SRCNN, while BN does not perform well on this task. The reason may
be that BN applies the same statistics to all patches of one image which
causes some overall intensity shift (see Figs. \ref{fig_single_SR_1}). From
the visual comparisons, we can see that our method not only enhances the
resolution but also removes artifacts, \eg, the ringing effect in Fig.
\ref{fig_single_SR_1}.



\subsection{Ablation studies and discussion}
Finally, we investigated the differential effects of the $\sigma^2$ term and
the L1 regularizer on the performance. We ran ablation studies on CIFAR-10/100
as well as PTB experiments. The results are listed in
Table~\ref{tab:ablation}.

We find that adding the smoothing term $\sigma^2$ and the L1 regularization
consistently increases the performance of the models. In the convolutional
networks, we find that L1 and $\sigma$ both have  similar effects on the
performance. L1 seems to be slightly more important.  In recurrent networks,
$\sigma^2$ has a much more dramatic effect on the  performance than the L1
regularizer.

\begin{figure}[t]
\begin{center}
\begin{tiny}

\resizebox{1.0\textwidth}{!}{
\begin{minipage}[t]{0.09\linewidth}
\centering
Baseline\\
\includegraphics[width=\linewidth]{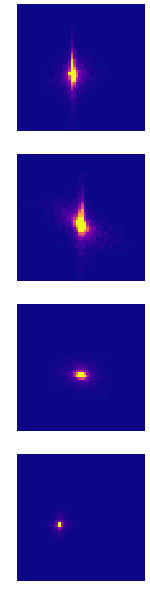}\\  
Corr. 0.19                                              \\  
MI 0.37                                                     
\end{minipage}

\begin{minipage}[t]{0.09\linewidth}
\centering
BN\\
\includegraphics[width=\linewidth]{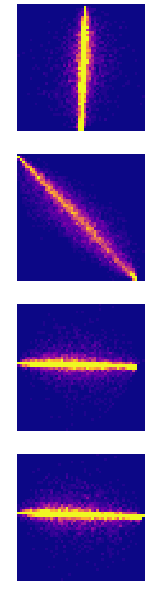}\\  
Corr. 0.43                                        \\  
MI 1.20                                               
\end{minipage}

\begin{minipage}[t]{0.09\linewidth}
\centering
BN +L1\\
\includegraphics[width=\linewidth]{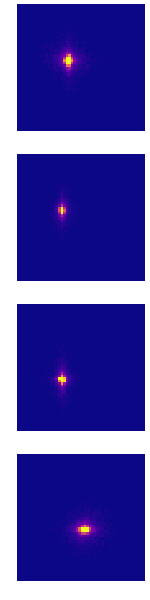}\\ 
Corr. 0.17                                          \\ 
MI 0.66                                                
\end{minipage}

\begin{minipage}[t]{0.09\linewidth}
\centering
BN-S\\
\includegraphics[width=\linewidth]{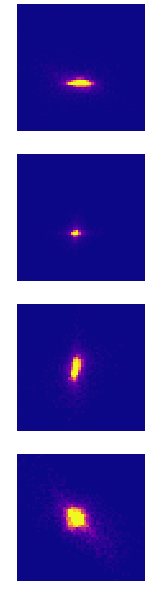} \\ 
Corr. 0.23                                          \\ 
MI 0.80                                                
\end{minipage}

\begin{minipage}[t]{0.09\linewidth}
\centering
BN*\\
\includegraphics[width=\linewidth]{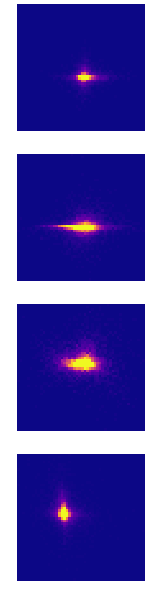} \\ 
Corr. 0.17                                          \\ 
MI 0.66                                                
\end{minipage}

\begin{minipage}[t]{0.09\linewidth}
\centering
LN\\
\includegraphics[width=\linewidth]{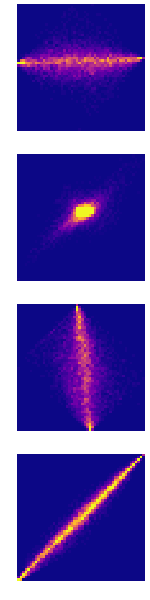}\\ 
Corr. 0.55                                        \\ 
MI 1.41                                              
\end{minipage}

\begin{minipage}[t]{0.09\linewidth}
\centering
LN +L1\\
\includegraphics[width=\linewidth]{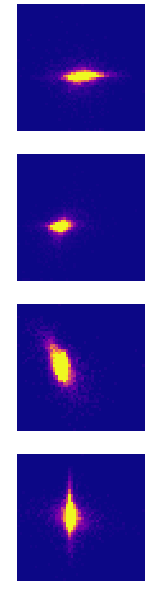} \\ 
Corr. 0.17                                           \\ 
MI 0.67                                                 
\end{minipage}

\begin{minipage}[t]{0.09\linewidth}
\centering
LN-S\\
\includegraphics[width=\linewidth]{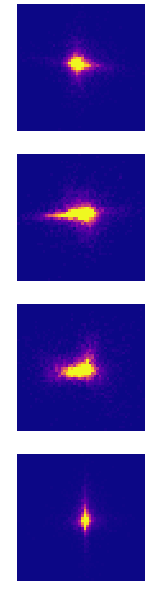} \\  
Corr. 0.20                                          \\  
MI 0.74                                                 
\end{minipage}

\begin{minipage}[t]{0.09\linewidth}
\centering
LN*\\
\includegraphics[width=\linewidth]{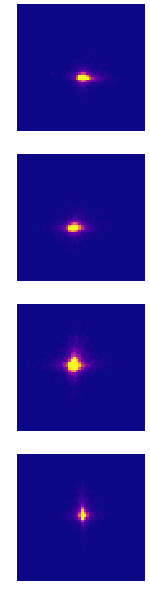} \\  
Corr. 0.16                                              \\  
MI 0.64                                                     
\end{minipage}

\begin{minipage}[t]{0.09\linewidth}
\centering
DN\\
\includegraphics[width=\linewidth]{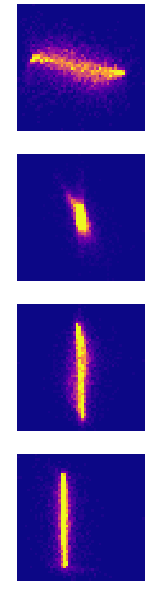} \\  
Corr. 0.21                                          \\  
MI 0.81                                                 
\end{minipage}

\begin{minipage}[t]{0.09\linewidth}
\centering
DN*\\
\includegraphics[width=\linewidth]{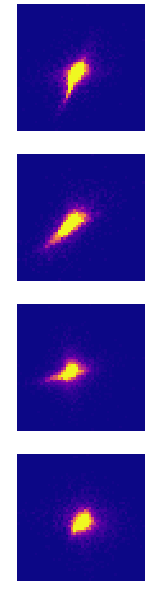} \\ 
Corr. 0.20                                          \\ 
MI 0.73                                                
\end{minipage}
}
\caption{First layer CNN pre-normalized activation joint histogram}
\label{fig:joint_histo}
\end{tiny}
\end{center}
\end{figure}

Fig.~\ref{fig:joint_histo} plots randomly sampled  pairwise pre-normalization
responses (after the linear transform) in the first layer at the same spatial
location of the feature map, along with the average pair-wise  correlation
coefficient (Corr) and mutual information (MI). It is evident that both
$\sigma$ and L1 encourages  independence of the learned linear filters.

There are several factors that could explain the improvement in performance.
As mentioned above, adding the L1 regularizer on the activations encourages
the filter responses to be less correlated. This can increase the robustness
of the variance estimate in the normalizer and lead to an improved scaling of
the responses to a good regime. Furthermore, adding the smoother to the
denominator in the normalizer can be seen as implicitly injecting zero mean
noise on the activations. While noise injection would not change the mean, it
does add a term to the variance of the data, which is represented
by $\sigma^2$. This term also makes the normalization equation invertible.
While dividing by the standard deviation decreases the degrees of freedom in
the data, the smoothed normalization equation is fully information preserving.
Finally, DN type operations have been shown to decrease the redundancy of
filter responses to natural images and sound \citep{Schwartz2001, Sinz2009,
Lyu2008a}. In combination with the L1 regularizer this could lead to a more
independent representation of the data and thereby increase the performance of
the network.

\begin{table}[t]
\caption{Comparison of standard batch and layer normalation (BN and LN)
  models, to those with only L1 regularizer (+L1), only the $\sigma$
  smoothing term (-s), and with both (*). We also compare divisive
  normalization with both (DN*), versus with only the smoothing term (DN).
}
\label{tab:ablation}
\begin{center}
\begin{tabular}{l|ll|lll}
{\bf Model} & {\bf CIFAR-10}&{\bf CIFAR-100}&{\bf LSTM}&{\bf Tanh RNN}&{\bf ReLU RNN}\\
\hline\hline                                                                                    
Baseline    &0.7565     &0.4409    & 115.720   & 149.357    & 147.630      \\ 
Baseline +L1&0.7839     &0.4517    & 111.885   & 143.965    & 148.572      \\
\hline                                                                             
BN          &0.7807     &0.4814    & 123.245   & 148.052    & 164.977      \\  
BN +L1      &0.8067     &0.5100    & 123.736   & 152.777    & 166.658      \\
BN-s        &0.8017     &0.5005    & 123.243   & 131.719    & 139.159      \\
BN*         &{\bf0.8179}&{\bf0.5156}& 116.920  & 129.155    & 138.947      \\
\hline                                                                       
LN          &0.7211     &0.4249    & 119.247   & 154.324    & 149.128      \\
LN +L1      &0.7994     &0.4990    & 116.964   & 152.100    & 147.937      \\
LN-s        &0.8083     &0.4863    & 102.492   & 133.812    & 118.786      \\
LN*         &0.8091     &0.4957    &{\bf101.725}& 129.823   & {\bf 116.609}\\
\hline                                                                          
DN          &0.8058     &0.4892    & 103.714   & 132.143     & 118.789     \\
DN*         &0.8122     &0.5066    & 102.238   & {\bf123.652}& 117.868     \\
\end{tabular}
\end{center}
\end{table}

\section{Conclusions} 

We have proposed a unified view of normalization techniques which contains batch and layer normalization as special cases. We have shown that when combined with a sparse regularizer on the activations, our framework has significant benefits over standard normalization techniques. We have demonstrated this in the context of  both convolutional neural nets as well as recurrent neural networks.
In the future we plan to explore other regularization techniques such as group
sparsity. We also plan to conduct a more in-depth analysis of the effects of
normalization on the correlations of the learned representations.

\newpage
\paragraph{Acknowledgements} RL is supported by Connaught International Scholarships. FS would like to thank Edgar Y. Walker, Shuang
Li, Andreas Tolias and Alex Ecker for helpful discussions. Supported by the Intelligence Advanced Research Projects Activity (IARPA) via Department of Interior/Interior Business Center (DoI/IBC) contract number D16PC00003. The
U.S. Government is authorized to reproduce and distribute reprints for
Governmental purposes notwithstanding any copyright annotation thereon.
Disclaimer: The views and conclusions contained herein are those of the
authors and should not be interpreted as necessarily representing the official
policies or endorsements, either expressed or implied, of IARPA, DoI/IBC, or
the U.S. Government.

\newpage
\bibliography{iclr2016_conference}
\bibliographystyle{iclr2016_conference}

\newpage
\appendix

\section{Effect of sigma and L1 on CIFAR-10/100 validation set}

\begin{figure}[h]
\centering

\begin{minipage}[t]{0.245\linewidth}
\centering
\includegraphics[width=\linewidth,trim={1cm 0cm 1cm 0.5cm},clip]{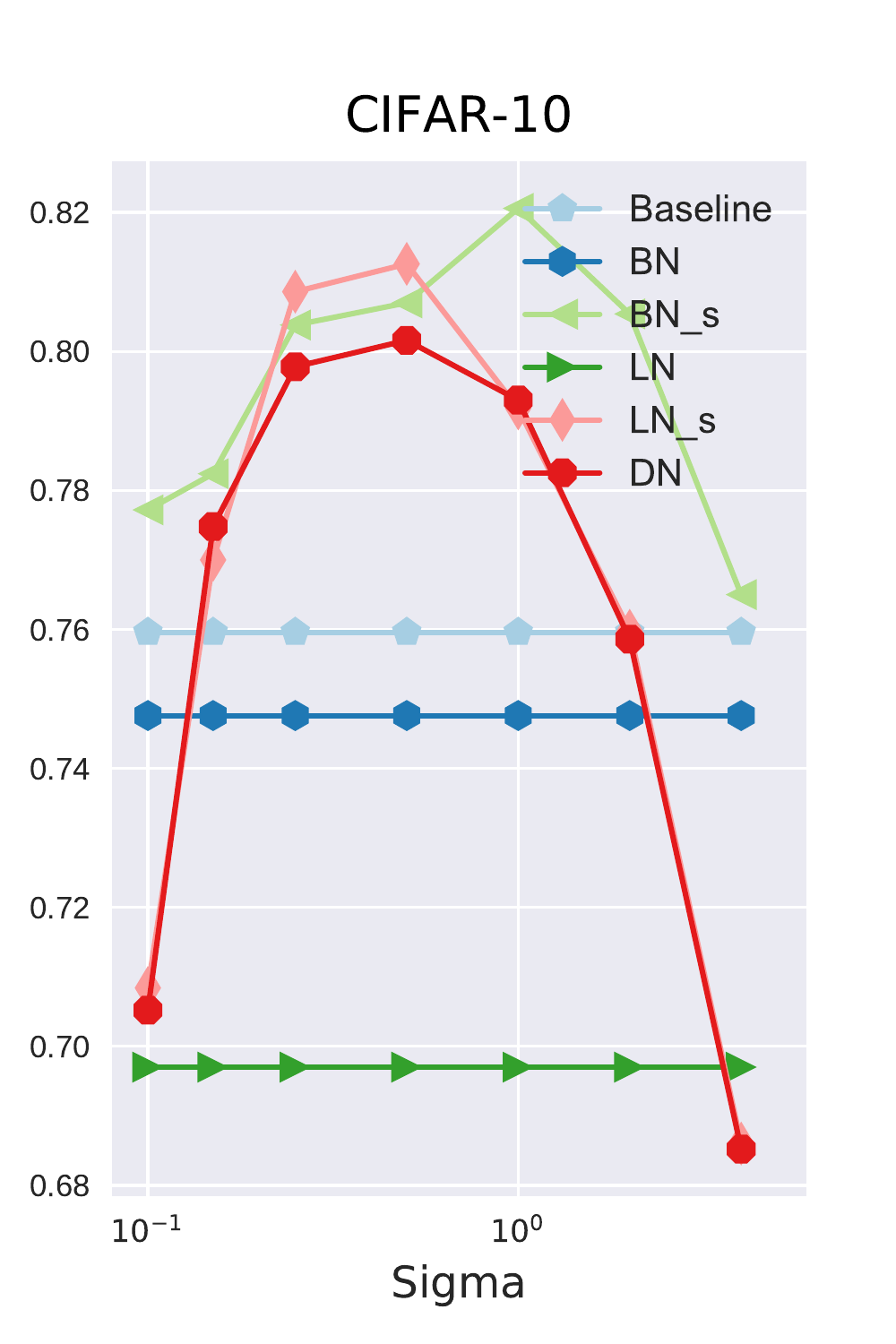}\\
(a)
\end{minipage}
\begin{minipage}[t]{0.245\linewidth}
\centering
\includegraphics[width=\linewidth,trim={1cm 0cm 1cm 0.5cm},clip]{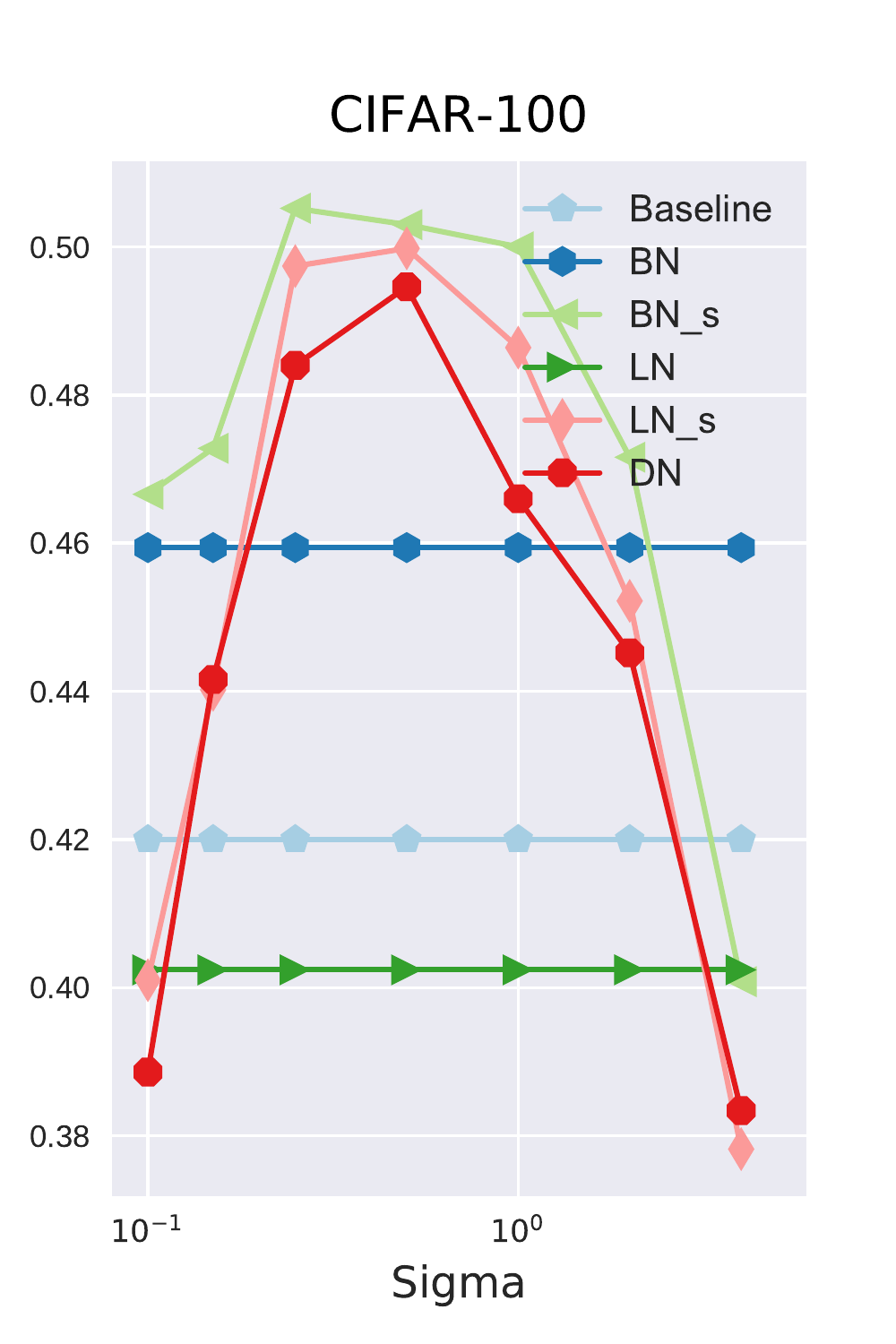}\\
(b)
\end{minipage}
\begin{minipage}[t]{0.245\linewidth}
\centering
\includegraphics[width=\linewidth,trim={1cm 0cm 1cm 0.5cm},clip]{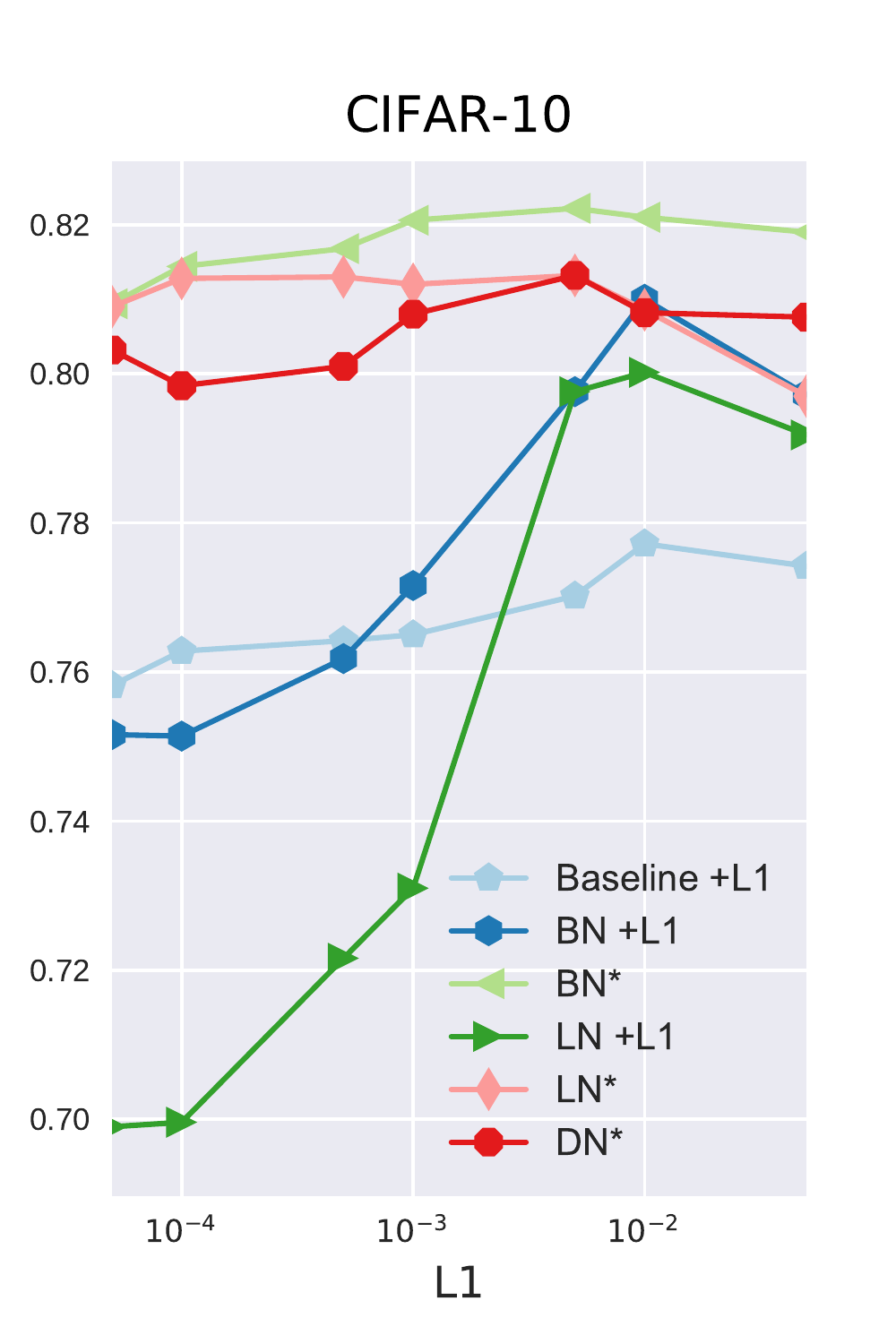}\\
(c)
\end{minipage}
\begin{minipage}[t]{0.245\linewidth}
\centering
\includegraphics[width=\linewidth,trim={1cm 0cm 1cm 0.5cm},clip]{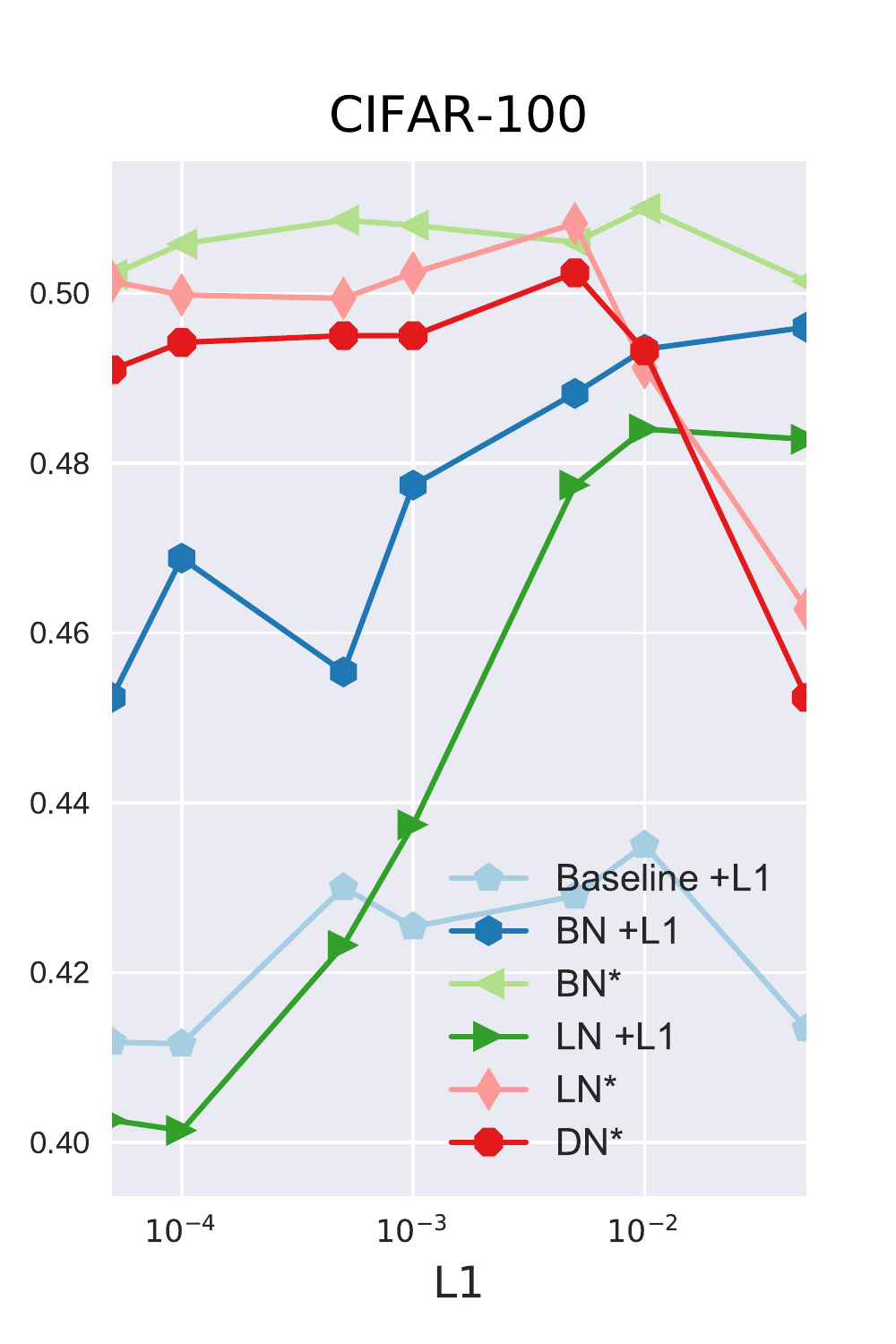}\\
(d)
\end{minipage}
\caption{Validation accuracy on CIFAR-10/100 showing effect of sigma constant (a, b) and L1 regularization (c, d) on BN, LN, and DN}
\label{fig:sigma}
\end{figure}

We plot the effect of $\sigma$ and L1 regularization on the validation
performance in Figure~\ref{fig:sigma}. While sigma makes the most
contributions to the improvement, L1 also provides much gain for the original
version of LN and BN.

\section{LSTM Implementation Details}

In LSTM experiments, we found that have an individual normalizer for each 
non-linearity (sigmoid and tanh) helps the performance for both LN and DN. 
Eq.~\ref{eq:lstm}-\ref{eq:lstm2} are the standard LSTM equations, and let $N$ 
be the normalizer function, our new normalizer is replacing the nonlinearity 
with Eq.~\ref{eq:new_norm}-\ref{eq:new_norm2}. 
This modification can also be thought 
as combining normalization and activation as a single activation function.

This is different from the released implementation of LN and BN in LSTM, which
separately normalized the concatenated vector $W_h \mathbf{h}_{t-1}$ and 
$W_x \mathbf{x}_t$. For all LN* and DN experiments we choose this new 
formulation, whereas LN experiments are consistent with the released version.
\begin{eqnarray}
\label{eq:lstm}
\left(\begin{array}{ccc}
\mathbf{f}_t \\
\mathbf{i}_t \\
\mathbf{o}_t \\
\mathbf{g}_t
\end{array}\right)
 &=&
 W_h \mathbf{h}_{t-1} +
 W_x \mathbf{x}_t +
 \mathbf{b}
 \\
\mathbf{c}_t &= &\sigma(\mathbf{f}_t) \odot \mathbf{c}_{t-1} +
\sigma(\mathbf{i}_t) \odot \tanh(\mathbf{g}_t) \\
\label{eq:lstm2}
\mathbf{h}_t &= &\sigma(\mathbf{o}_t) \odot \tanh(\mathbf{c}_t)
\end{eqnarray}
\begin{eqnarray}
\label{eq:new_norm}
\bar{\sigma}(x) &= & \sigma(N(x))\\
\label{eq:new_norm2}
\overline{\tanh}(x) &= & \tanh(N(x))
\end{eqnarray}

\section{More results on Image Super-Resolution}
\begin{figure}[h]
\begin{minipage}[t]{0.24\linewidth}
\centering
\includegraphics[width=1.0\linewidth]{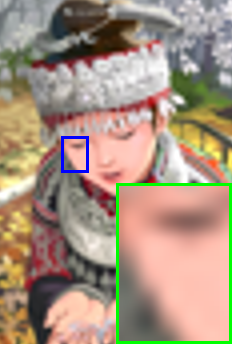}
{PSNR 21.69dB}
\end{minipage}
\begin{minipage}[t]{0.24\linewidth}
\centering
\includegraphics[width=1.0\linewidth]{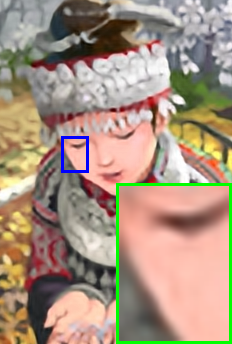}
{PSNR 22.62dB}
\end{minipage}
\begin{minipage}[t]{0.24\linewidth}
\centering
\includegraphics[width=1.0\linewidth]{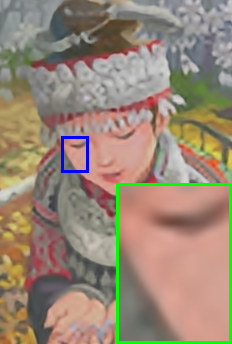}
{PSNR 20.06dB}
\end{minipage}
\begin{minipage}[t]{0.24\linewidth}
\centering
\includegraphics[width=1.0\linewidth]{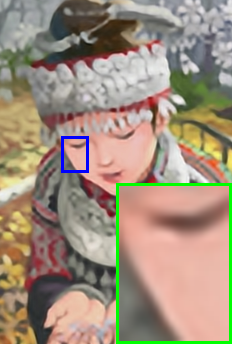}
{PSNR 22.69dB}
\end{minipage} \\ 
\begin{minipage}[t]{0.24\linewidth}
\centering
\includegraphics[width=1.0\linewidth]{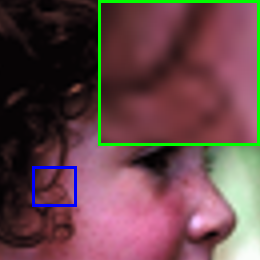}
{PSNR 31.55dB \\ (a) Bicubic}
\end{minipage}
\begin{minipage}[t]{0.24\linewidth}
\centering
\includegraphics[width=1.0\linewidth]{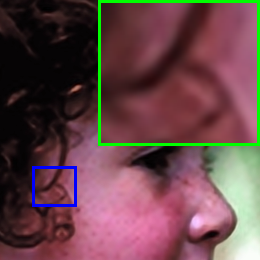}
{PSNR 32.29dB \\ (b) SRCNN}
\end{minipage}
\begin{minipage}[t]{0.24\linewidth}
\centering
\includegraphics[width=1.0\linewidth]{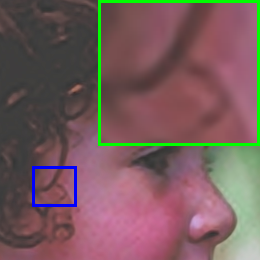}
{PSNR 19.39dB \\ (c) BN}
\end{minipage}
\begin{minipage}[t]{0.24\linewidth}
\centering
\includegraphics[width=1.0\linewidth]{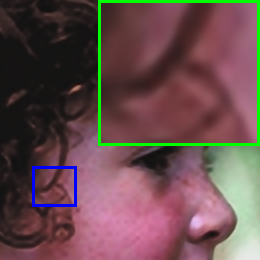}
{PSNR 32.31dB \\ (d) DN*}
\end{minipage}\vspace{0.1in}
\caption{Comparisons at a magnification factor of 4.}
\label{fig_single_SR_3}
\end{figure}

We include results on another standard dataset Set5 \cite{bevilacqua2012low} in
Table \ref{tbl_sr_3} and show more visual results in Fig. 
\ref{fig_single_SR_3}.

\begin{table}[h]
\begin{center}
\caption{Average test results of PSNR and SSIM on Set5 Dataset.}
\begin{tabular}{c | c c c c }
\textbf{Model} & \textbf{PSNR} (x3) & \textbf{SSIM} (x3) & \textbf{PSNR} (x4) & \textbf{SSIM} (x4) \\
\hline \hline
Bicubic		& 30.41 & 0.8678 & 28.44 & 0.8097 \\
A+			& 32.59 & 0.9088 & 30.28 & 0.8603 \\
SRCNN     & \textbf{32.83} & 0.9087 & 30.52 & 0.8621 \\ 
BN   			& 22.85 & 0.8027 & 20.71 & 0.7623 \\
DN* 			& \textbf{32.83} & \textbf{0.9106} & \textbf{30.62} & \textbf{0.8665}  \\
\end{tabular}
\label{tbl_sr_3}
\end{center}
\end{table}

\end{document}